\documentclass[conference]{IEEEtran}
\IEEEoverridecommandlockouts

\usepackage{cite}
\usepackage{amsmath,amssymb,amsfonts}
\usepackage{algorithmic}
\usepackage{graphicx}
\usepackage{textcomp}
\usepackage{xcolor}
\usepackage{hyperref}
\usepackage{capt-of}
\usepackage[ruled,vlined]{algorithm2e}
\usepackage{booktabs}
\usepackage{multirow}

\newcommand{\Efuse}{\texorpdfstring{$E_{fuse}$}{Efuse}}

\begin{document}

\title{
Re-engineering SORT-based algorithms for low-cost small object tracking from omnidirectional footage
}
\author{
\IEEEauthorblockN{
Xin Shu\IEEEauthorrefmark{1},
Meegan Gower\IEEEauthorrefmark{1},
Yvonne Buckley\IEEEauthorrefmark{2},
Anil Kokaram\IEEEauthorrefmark{1}
}

\IEEEauthorblockA{
\IEEEauthorrefmark{1}Sigmedia Group, Electronic and Electrical Engineering, Trinity College Dublin
}

\IEEEauthorblockA{
\IEEEauthorrefmark{2}School of Zoology, Trinity College Dublin
}

\IEEEauthorblockA{
\texttt{\{xins,gowerm,buckleyy,anil.kokaram\}@tcd.ie}
}

\thanks{Supported by the project \textit{Digitising Biodiversity} sponsored by grant from the Kinsella Foundation to Trinity College Dublin.}
}

\maketitle

\begin{abstract}
Multi-object tracking (MOT) has advanced rapidly in urban surveillance and autonomous driving, yet many trackers rely on ReID- and transformer-based appearance encoders and are designed for standard FoV cameras. These assumptions break down for low-cost omnidirectional deployments, where equirectangular projection introduces seam discontinuities and targets appear to be small and fast-moving. We address multi-object tracking of flying animals captured in remote environments using omnidirectional cameras. We propose a lightweight framework that re-engineers SORT-based tracking for this geometry, including (i) a Seam-Aware Motion Model that keeps the Kalman state continuous across the seam, (ii) a composite seam-aware association cost that pairs a wrapped Euclidean term with GIoU, and (iii) OmniSmall, a new benchmark of omnidirectional wildlife footage. On our new dataset, with ground-truth detections, our modifications improved over OCSORT by +8.51 HOTA, +9.41 MOTA, and +10.17 IDF1; with YOLOX detections the gain narrows to +1.95 HOTA. Our proposed methods improved tracking performance on OmniSmall and remained competitive on JRDB without adding appearance encoders while keeping the tracking stage CPU-only. Our dataset and source code are available at: \href{https://github.com/Xin-Shu/OmniSORT.git}{\underline{https://github.com/Xin-Shu/OmniSORT.git}}.
\end{abstract}
\begin{IEEEkeywords}
multi-object tracking, omnidirectional video, small object tracking, track-by-detection, wildlife monitoring
\end{IEEEkeywords}

\section{Introduction}
\label{sec:intro}
Multi-object tracking (MOT) plays a central role in many computer vision applications, with growing interest in online and real-time trackers. Algorithms such as SORT~\cite{sort-2016} and the variants~\cite{deepsort-2017,botsort_2022,strongsort-2023,ocsort-2023,hybridrsort_2024} have shown strong performance in structured, urban scenes. However, their application to low-cost, remote deployments, such as monitoring wild animals using omnidirectional cameras, remains underexplored. These scenarios are marked by non-Euclidean image geometry, tiny object scale, and strict power and compute constraints, which together inhibit deployments of many conventional MOT pipelines.

Our work targets small-object tracking in omnidirectional footage, a regime that differs from standard field-of-view (FoV) cameras in both geometry and target scale. The equirectangular geometry typical of omnidirectional footage causes object motion to appear non-linear when it is not, and crossing seam boundaries causes discontinuous trajectories. These issues violate the linear assumptions used in standard trackers.

\begin{figure*}
    \centering
    \includegraphics[width=0.49\linewidth]{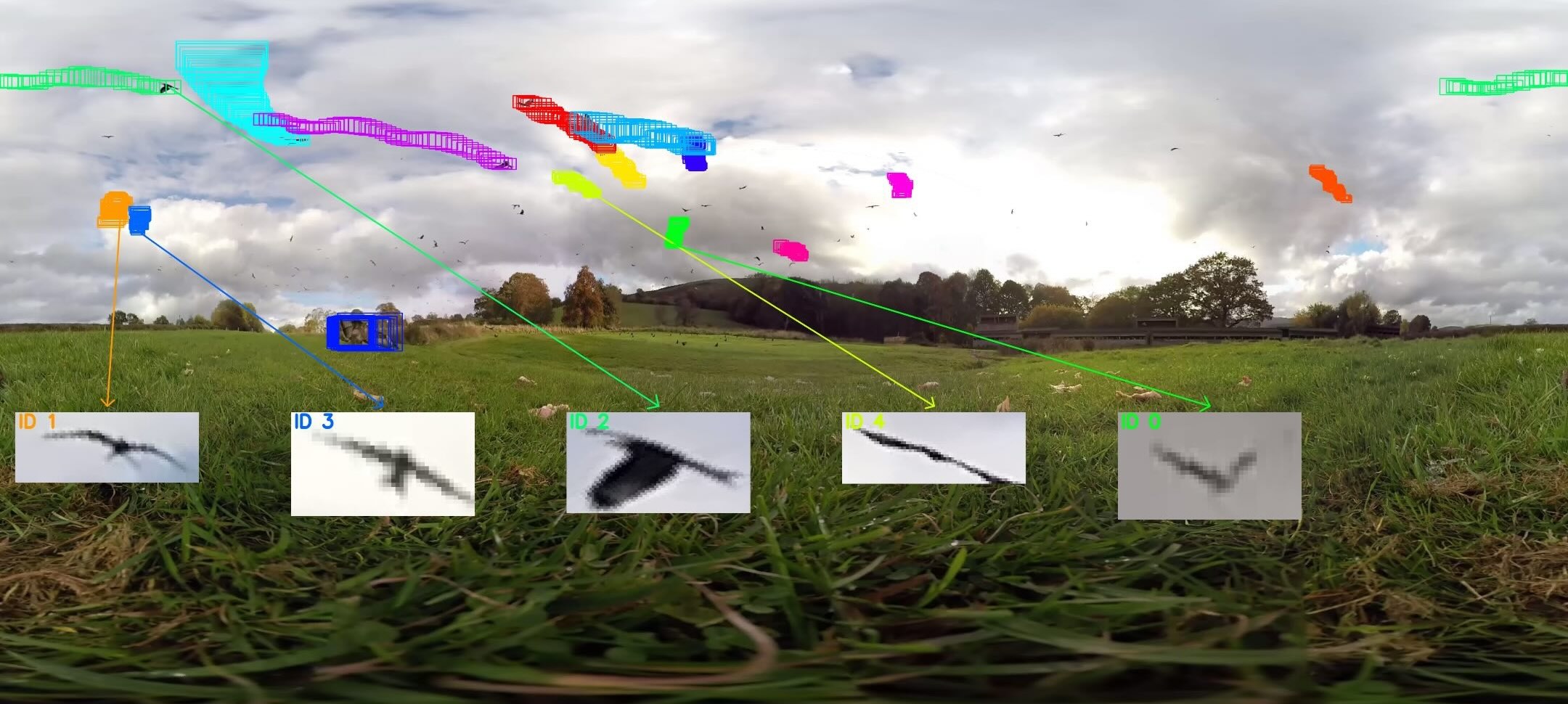}
    \includegraphics[width=0.49\linewidth]{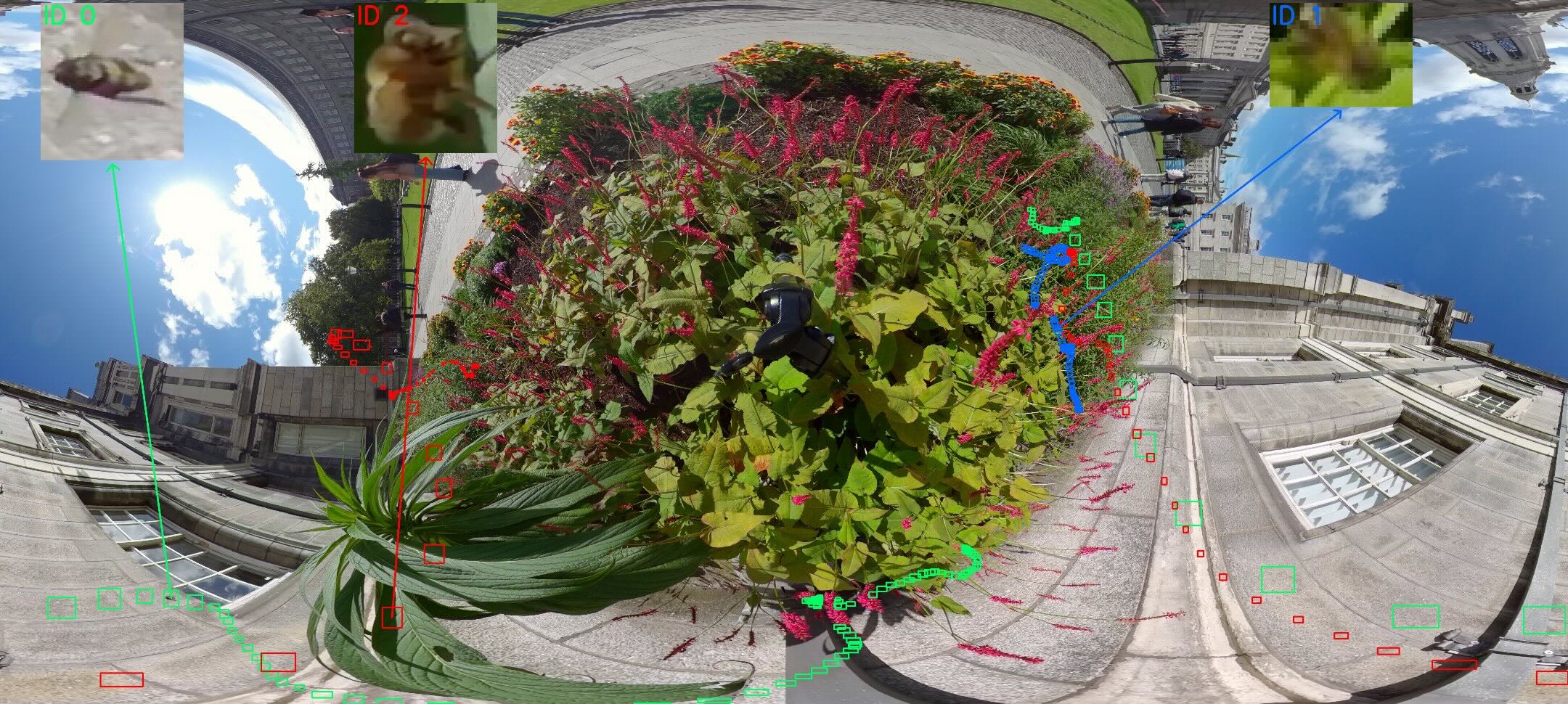}\\
    \vspace{0.1cm}
    \includegraphics[width=0.49\linewidth]{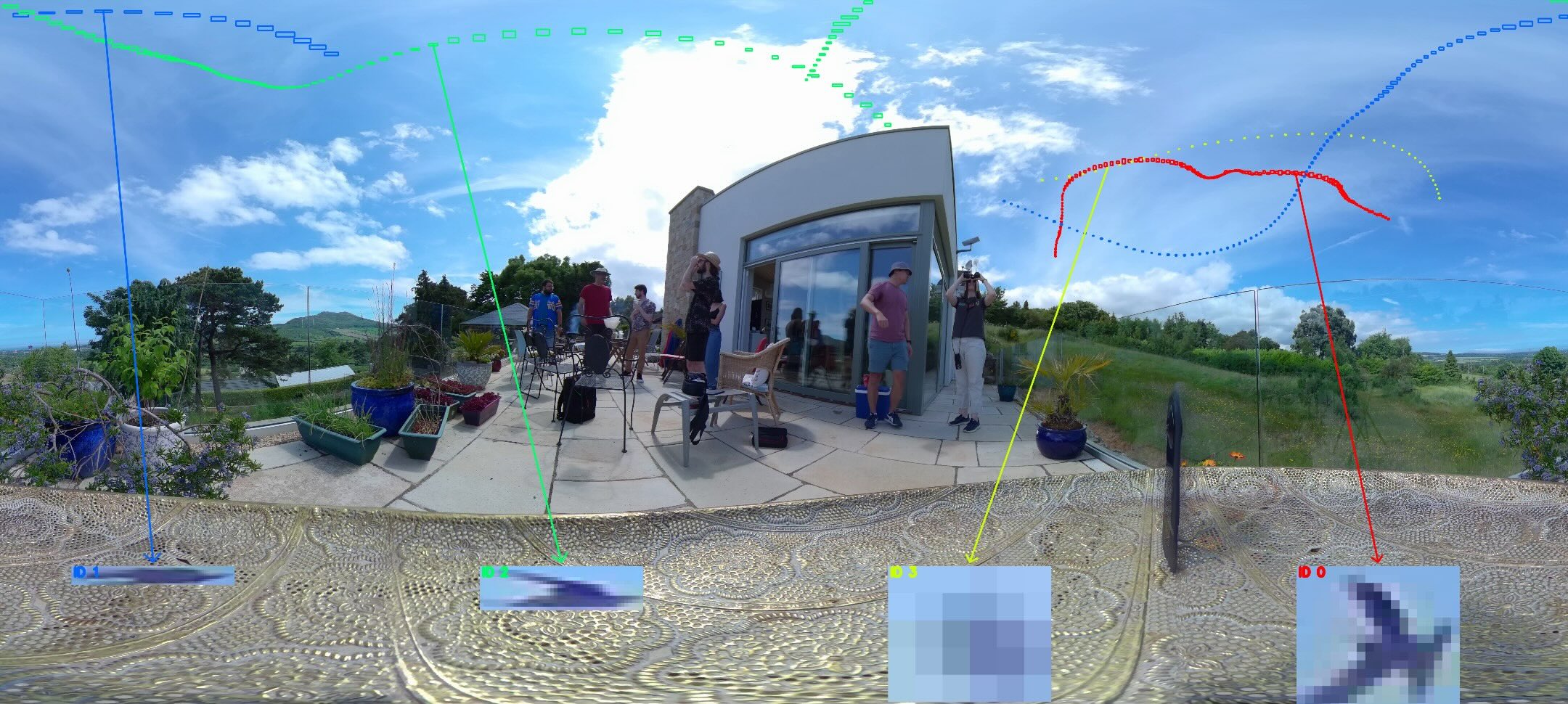}
    \includegraphics[width=0.49\linewidth]{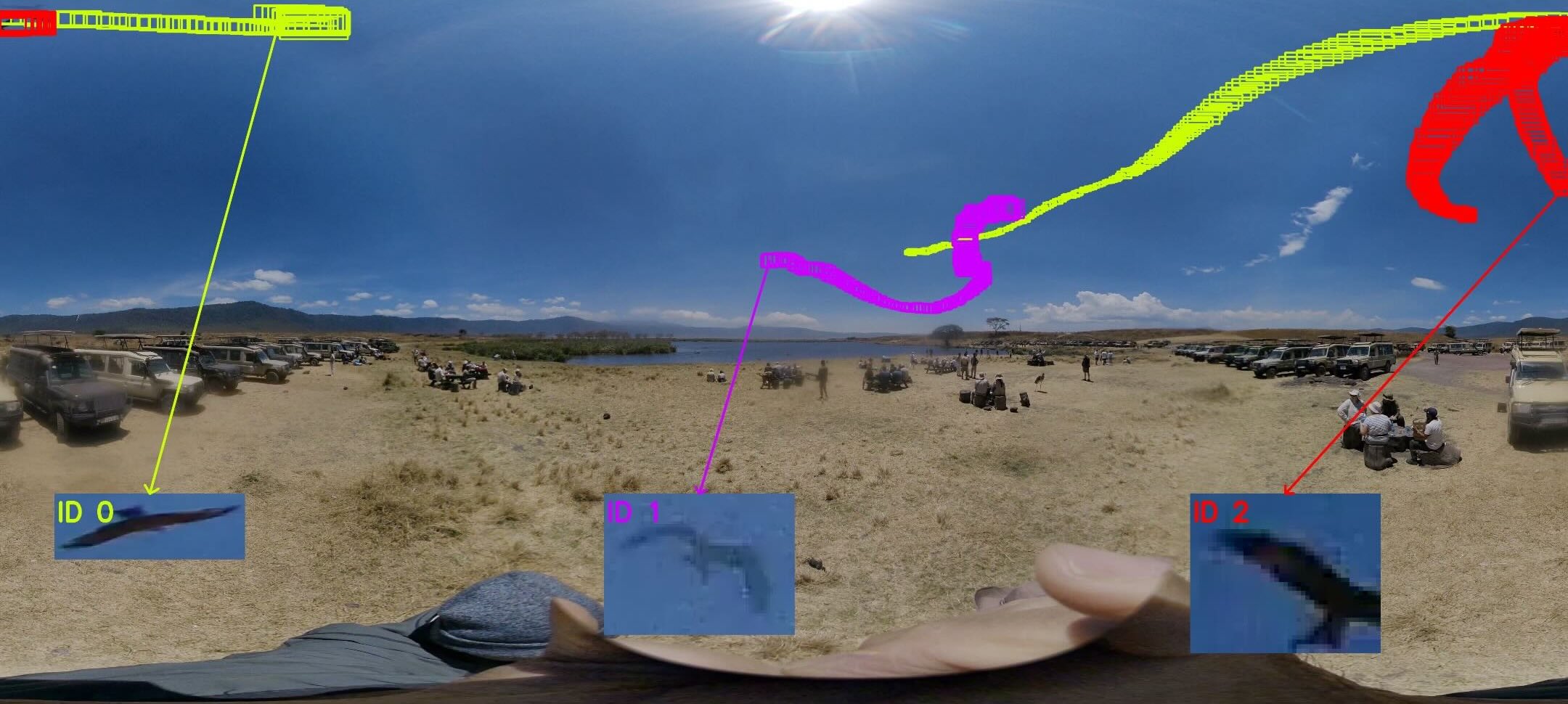}\\
    \vspace{-.7em}
    \caption{
    Representative frames from four omnidirectional sequences illustrating object trajectories. Coloured bounding boxes accumulated over time visualise the per-object tracks in the equirectangular projection. The \textbf{upper-right frame} shows honey bees around a flower bed (after a $90^{\circ}$ vertical rotation of the projection), while the \textbf{remaining frames} show avian species. Insets show zoomed-in crops of several tracked identities, highlighting strong appearance ambiguity.}
    \vspace{-1.6em}
    \label{fig:small_mot}
\end{figure*}

Small objects (typically below $32\times32$ pixels~\cite{tiny_review_2024}) in such footage also dramatically reduce the effectiveness of appearance models used for matching or association in many SOTA trackers. A rectangular box around a flying target is mostly background at any scale, and at these sizes the few foreground pixels left carry too little texture for a learned embedding to tell identities apart. The insets in Fig.~\ref{fig:small_mot} show that at this small scale, most of the box area is occupied by background regardless of motion. Hence appearance models are less discriminative because the matching score will be dominated by background. Furthermore, limited connectivity and power make it impractical to rely on ReID- and transformer-based appearance encoders~\cite{deepreid_2014,deepsort-2017,transreid_2021}. SORT and similar motion-based trackers are attractive due to their simplicity and low resource demands, but their reliance on smooth motion and IoU-based matching leads to frequent identity switches in our setting.

In this paper, we re-engineer two SORT-based tracking frameworks~\cite{sort-2016,ocsort-2023} to address these challenges while preserving a lightweight design. Our contributions are as follows: (1) we introduce a Seam-Aware Motion Model (SAMM) that maintains horizontal continuity across the left-right equirectangular seam; (2) we propose \Efuse, a cost that combines \textit{Omni-Euclidean} (OmniEuc) distance with \textit{Generalised Intersection over Union} (GIoU)~\cite{giou_2019} for robust object association in omnidirectional footage; (3) we present \textit{OmniSmall}, a new benchmark for small-object tracking in omnidirectional footage capturing real-world trajectories under severe distortion and left-right seam crossings.

Our proposed method outperformed the SORT and OCSORT baselines on the OmniSmall dataset and achieved competitive performance on the JRDB dataset~\cite{jrdb-2023}, while adding no appearance model and no GPU dependency at the tracking stage.

\vspace{-.7em}
\section{Background}
\label{sec:background}
The tracking-by-detection (TBD) paradigm remains dominant in MOT, where detectors provide per-frame object locations and trackers maintain temporal identity consistency. Simple Online and Realtime Tracking (SORT)~\cite{sort-2016} introduced a widely adopted approach using a Kalman filter~\cite{kalman-1960} for prediction and the Hungarian algorithm~\cite{hungarian-1955} for frame-to-frame data association, with Intersection over Union (IoU) as the association cost. Subsequent works have refined the motion model~\cite{bytetrack-2022,strongsort-2023,ocsort-2023,hybridrsort_2024}, introduced learned appearance cues~\cite{deepsort-2017,botsort_2022,deepocsort_2023}, or adopted stronger association strategies~\cite{bytetrack-2022,strongsort-2023,hybridrsort_2024}. These extensions have been validated on urban surveillance and autonomous driving benchmarks, where targets are pedestrian- or vehicle-scale, motion is locally linear in image coordinates, and re-identification features are discriminative. All three assumptions break for our deployment regime: small targets carry little appearance signal because the bounding box is dominated by background; flying-animal motion in equirectangular projection is non-linear and discontinuous at the seam; and ReID- or transformer-based appearance encoders~\cite{deepreid_2014,reid_2018,person_reid_2019,transreid_2021} exceed the compute budget of a remote, battery-powered camera. SORT itself remains attractive in this setting precisely because it makes none of these assumptions about appearance, but the IoU-based association and the linear Kalman prediction fail when targets are small, fast, and projected onto a wrapped image plane.

Three bodies of work address different aspects of our problem. Firstly, omnidirectional tracking itself has attracted growing interest as a single camera can capture the full surroundings. Benchmarks, such as CVIP360~\cite{cvip360_2021} and 360VOT~\cite{360vot_2023}, characterise how equirectangular distortion and seam discontinuities affect tracking; both focus, however, on conventional targets in controlled surveillance scenes (CVIP360) or single-object protocols (360VOT) rather than our problem of multi-object tracking of small, fast targets. The most related learning-based work, OmniTrack~\cite{omnitrack-2025}, adapts MOT to omnidirectional imagery via panorama-aware feature learning and appearance association, achieving good performance on human-scale targets while inheriting the compute and appearance assumptions that our work is explicitly designed to avoid. Secondly, work in small-object tracking~\cite{tiny_review_2024} addresses targets below $32\times32$ pixels that render IoU-based metrics and appearance encoders unreliable. In that work, motion-only association becomes the dominant signal, but only for images from standard FoV cameras. Thirdly, work in wildlife and ecological MOT~\cite{animaltrack_2023,bucktales_2024} has driven practical demand for low-cost, deployable tracking but has not yet engaged with omnidirectional capture. The intersection of these three regimes (small targets, omnidirectional projection, and a strict on-device compute envelope) is where SORT-style trackers would be most useful and is the gap this paper addresses.

In TBD, the association cost is central to linking predicted box locations with detections. IoU becomes unreliable when boxes do not overlap, and several extensions have been proposed (e.g., DIoU and CIoU~\cite{diou_2020}). We adopt Generalised IoU (GIoU)~\cite{giou_2019}, which extends IoU with a penalty based on the smallest enclosing region $C_{mn}$ of two boxes $b_m,b_n$, so that non-overlapping boxes are still ordered by separation:
\begin{align}
    \text{IoU}_{(m,n)} &= \frac{|b_m \cap b_n|}{|b_m \cup b_n|}, \in [0,1] \\
    \text{GIoU}_{(m,n)} &= \text{IoU}_{(m,n)} - \frac{|C_{mn}| - |b_m \cup b_n|}{|C_{mn}|}, \in (-1,1]
    \label{eq:giou}
\end{align}
Following standard MOT practice, we report HOTA~\cite{hota_2021}, MOTA~\cite{mota_2008}, IDF1~\cite{idf1_2016}, and IDSw~\cite{mota_2008}: together these capture detection quality, association quality, identity preservation, and switch frequency, with HOTA as the primary measure since it jointly weights detection and association.

\section{Proposed Methods}
\label{sec:algorithms}
Our modifications to SORT~\cite{sort-2016} and OCSORT~\cite{ocsort-2023} focus on enhancing data association and motion modelling to better handle the geometric distortions and motion characteristics unique to tracking small objects from omnidirectional footage.

\subsection{SAMM: Seam-Aware Motion Model}
Object motion is modelled with a Kalman filter (\textit{Kalman unit}) using the SORT~\cite{sort-2016} constant-velocity formulation, extended with an aspect-ratio derivative ($\dot{r}$), giving the state vector $k_i=[u, v, s, r, \dot{u}, \dot{v}, \dot{s}, \dot{r}]$: $(u,v)$ denote the centre coordinates of the bounding box, $s$ is the scale (area), $r$ is the aspect ratio, and $\dot{u}, \dot{v}, \dot{s}, \dot{r}$ are the corresponding derivatives. In omnidirectional imagery, objects may cross the image seam (i.e., the left–right boundary of an equirectangular projection), causing an apparent discontinuity in coordinates (e.g., the green trajectory in the lower-right image in Fig.~\ref{fig:small_mot}). This disrupts the velocity estimation of the Kalman filter. We introduce a seam-aware correction: when the relative change in the horizontal velocity, $|\dot{u}_{f+1}-\dot{u}_{f}|/\max(|\dot{u}_{f}|,\epsilon)$, exceeds a threshold $\tau$, we interpret it as a seam crossing and wrap the updated horizontal velocity onto the signed principal interval $[-\tfrac{1}{2},\tfrac{1}{2})$ to undo the seam-induced velocity jump~\cite{markovic2016}, as described in Algorithm~\ref{algo:seam-crossing}. This keeps the horizontal component of the motion model continuous across the equirectangular seam. In this paper we set $\tau=10.0$ and $\epsilon=10^{-6}$.

\vspace{-.5em}
\begin{algorithm}
\KwIn{\\
\hspace{0.5 cm}Observed boxes at frame $f+1$:$\{B_{(f+1,i)}^{\text{ob}}\}_{i=1}^N$, \\
\hspace{1.0 cm}$b \gets [u,v,w,h]$ normalised in $[0,1]$,\\
\hspace{0.5 cm}Tracked Kalman unit at frame $f$: $\{K_{(f,j)}\}_{j=1}^{J}$,\\
\hspace{1.0 cm}$k \gets [u,v,s,r,\dot{u},\dot{v},\dot{s},\dot{r}]$,\\
\hspace{0.5 cm}Velocity-change ratio threshold: $\tau$
}
\KwOut{Tracked Kalman unit at frame $f+1$: $\{K_l\}_{l=1}^{L}$}
\nl Seam-aware Kalman prediction (position):\\
\hspace{0.4cm}$\hat{k}_j \gets [\hat{u},\hat{v},\hat{s},\hat{r},\hat{\dot{u}},\hat{\dot{v}},\hat{\dot{s}},\hat{\dot{r}}]\gets k_j.$predict$()$, \\
\hspace{0.4cm}$\hat{u} = \hat{u} \bmod 1$\\
\nl Update tracked Kalman unit (speed): \\
Associate $b_j$ to tracked Kalman unit $k_j$ at frame $f+1$\\
\hspace{0.4cm}$k_{(f,j)} \gets [u,v,s,r,\dot{u},\dot{v},\dot{s},\dot{r}]_{f}$, \\
\hspace{0.4cm}$b_{(f+1,j)} \gets [u,v,w,h]_{f+1}$ \\
\hspace{0.4cm}$k_{(f+1,j)} \gets  k_{(f,j)}$.update($b_{(f+1,j)}$)\\
Velocity correction:\\
\hspace{0.4cm}if $(|\dot{u}_{f+1}-\dot{u}_{f}|\ /\ \max(|\dot{u}_{f}|, \epsilon))>\tau$: \\
\hspace{0.8cm}$\dot{u}_{f+1} = \bigl((\dot{u}_{f+1} + \tfrac{1}{2}) \bmod 1\bigr) - \tfrac{1}{2}$\\
\nl Return: $\{K_l\}_{l=1}^{L}$.
\caption{{\bf SAMM} \label{algo:seam-crossing}}
\end{algorithm}

\vspace{-1.2em}
\subsection{OmniEuc: Seam-Aware Euclidean Distance}
In small-object tracking for omnidirectional views, IoU-based association can be unreliable. Small objects occupy only a few pixels but often move quickly, so Kalman-predicted locations are more likely to have no intersection with the observations at the next frame. Therefore, we propose to replace IoU with a Euclidean distance as the primary association cost for small-object tracking.

We define an \emph{Omni-Euclidean} (OmniEuc) distance as the main seam-aware association cost for omnidirectional tracking. In equirectangular images, targets may wrap across the left–right seam, causing apparent jumps that inflate standard Euclidean distances in image coordinates. To address this, we normalise box centres to $[0,1]$ and compute the minimum between the direct centre distance and the wrapped counterpart obtained by shifting by one image width, as formalised in Algorithm~\ref{algo:omni_euc}.
This yields a robust association cost that remains small for seam-crossing trajectories and large for unrelated detections.

\vspace{-.5em}
\begin{algorithm}
\KwIn{
    Observed boxes at frame $f$: $\{B_i^{\text{ob}}\}_{i=1}^N$, \\
    \hspace{1.2cm}Predicted boxes at frame $f$: $\{B_j^{\text{pr}}\}_{j=1}^M$, \\
    \hspace{1.7cm}$b \gets [u,v,w,h]$ normalised in $[0,1]$ \\
}
\KwOut{Omni-Distance matrix $E \in \mathbb{R}^{M \times N}$}
\nl Denote centre locations of boxes as: \\
\hspace{0.5cm}$(u_i^{\text{ob}},v_i^{\text{ob}})$,$(u_j^{\text{pr}},v_j^{\text{pr}})$;\\
\nl Compute standard Euclidean distance: \\
\hspace{0.5cm}$d_{ij} \gets \sqrt{(u_i^{\text{ob}} - u_j^{\text{pr}})^2  + (v_i^{\text{ob}} - v_j^{\text{pr}})^2}$; \\
\nl Compute wrapped Euclidean distance: \\
\hspace{0.5cm}$\Delta u=|u_i^{\text{ob}} - u_j^{\text{pr}}|,  \Delta v=|v_i^{\text{ob}} - v_j^{\text{pr}}|$ \\
\hspace{0.5cm}$\Delta \tilde{u}=\min(\Delta u, 1-\Delta u), $\\
\hspace{0.5cm}$\tilde{d}_{ij} \gets \sqrt{(\Delta \tilde{u})^2+(\Delta v)^2}$\\
\nl Return:\\
\hspace{0.5cm}$E_{ij} = \frac{\sqrt{2}}{2} \times \min(d_{ij},\tilde{d}_{ij})$,\\
\hspace{0.5cm}$E_{ij}\in[0, 1].$
\caption{{\bf OmniEuc} \label{algo:omni_euc}}
\end{algorithm}

\vspace{-1.2em}
\subsection{\Efuse: Composite Association Cost}
\begin{figure*}[ht]
    \centering
    \includegraphics[width=0.70\linewidth]{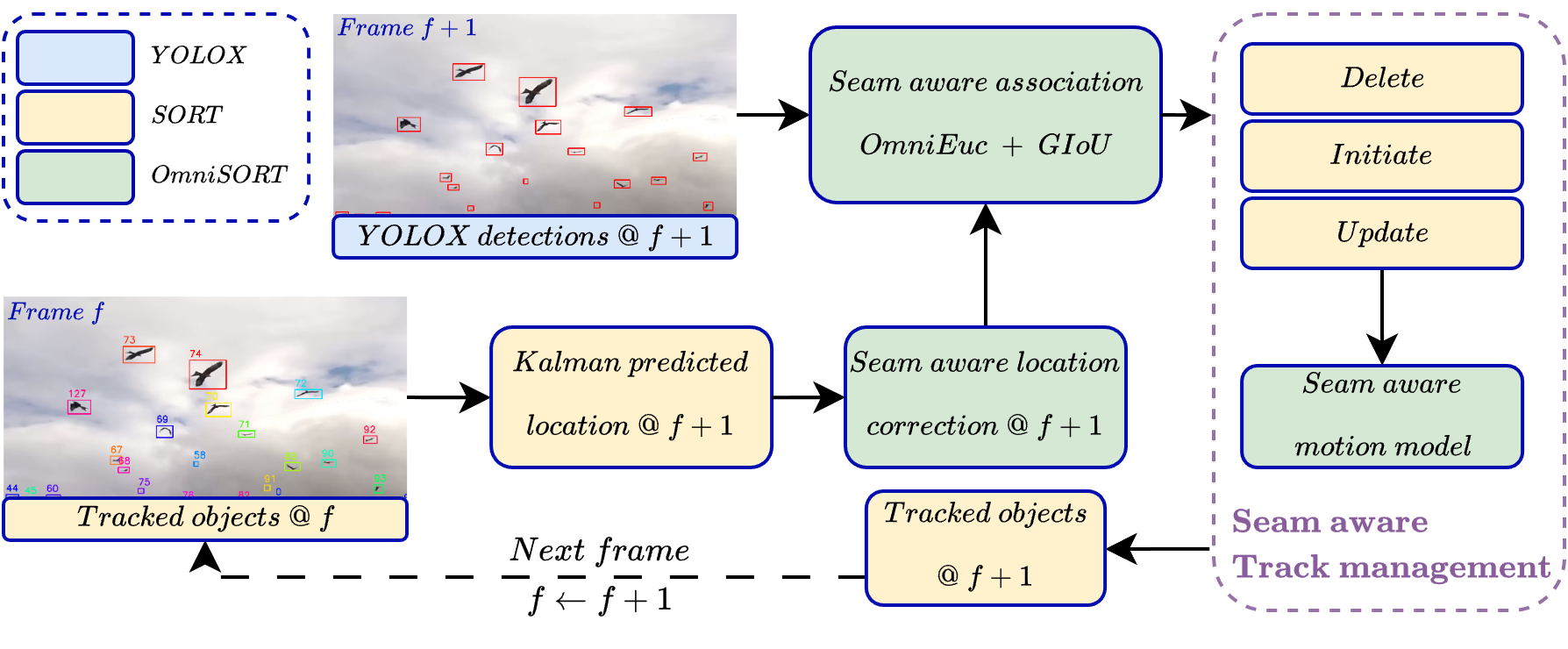}
    \includegraphics[width=0.29\linewidth]{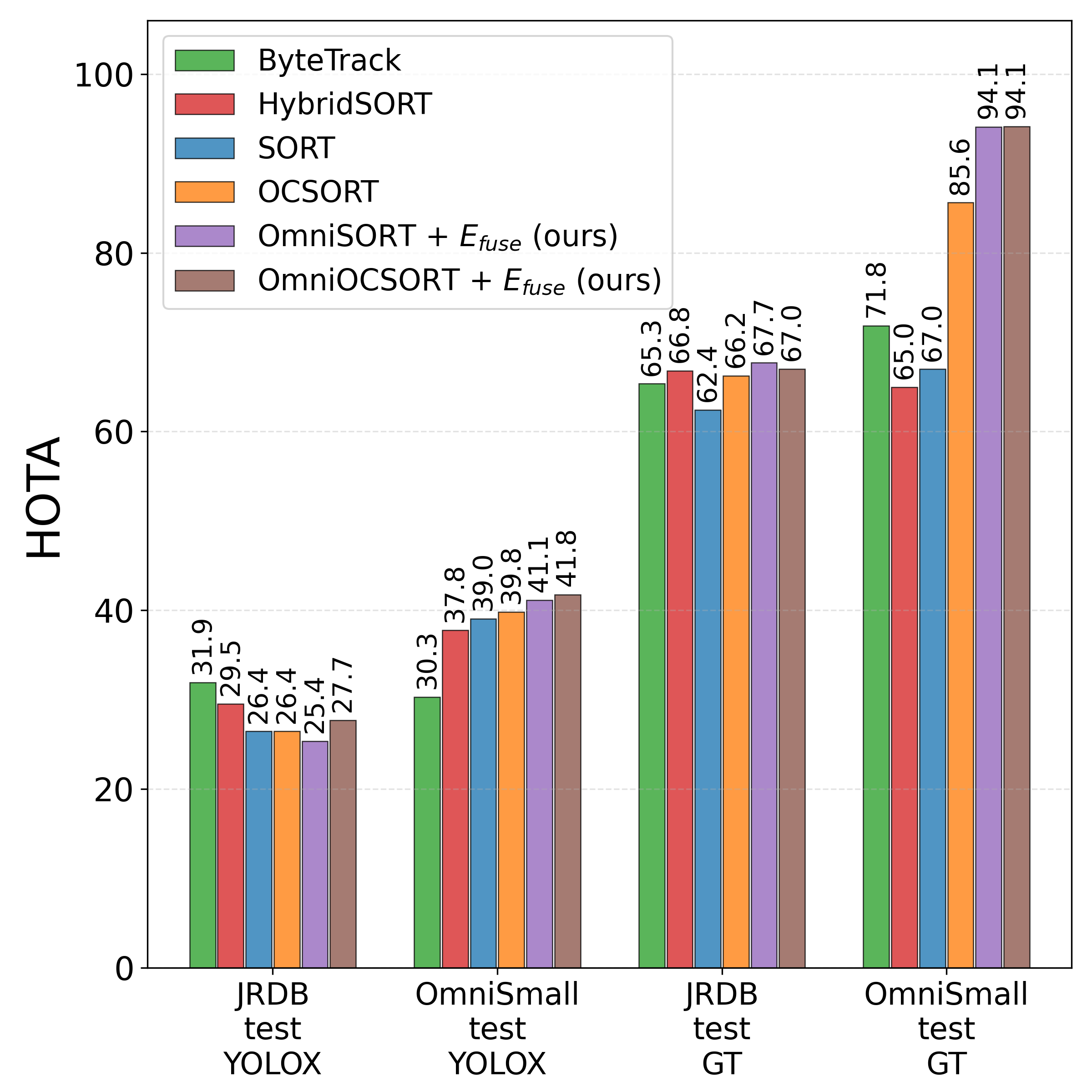}
    \vspace{-1.5em}
    \caption{
    Proposed OmniSORT pipeline (\textbf{left}), highlighting standard SORT components and the seam-aware modifications introduced in this work. HOTA comparison on JRDB and OmniSmall under ground-truth (GT) and YOLOX detections (\textbf{right}). Each group reports the performance of the evaluated tracking methods under one dataset and detection setting.}
    \label{fig:result_plot}
    \vspace{-1.3em}
\end{figure*}

Euclidean distance and IoU-based cost capture different aspects of similarity. Euclidean distance (in our case, OmniEuc) remains informative even when boxes do not overlap or lie across the left-right seams, while IoU and GIoU~\cite{giou_2019} are more useful when detections and predictions are well aligned and the patches overlap. In our setting, small objects and non-linear motions often cause Kalman-predicted locations to have little or no overlap with the detected observations in the next frame, so IoU-based costs become unreliable for identifying correct associations. To exploit the complementary properties of Euclidean and IoU measures, we define a composite association cost ($E_{fuse} \in [0,1]$) that employs a weighted combination of the normalised OmniEuc and GIoU-derived costs as follows.
\begin{equation}
\begin{aligned}
    E_{GIoU} &= \frac{1}{2} \times (1 - \mathrm{GIoU})\\
    E_{fuse} &= \lambda E_{OmniEuc} + (1-\lambda) E_{GIoU}
\end{aligned}
\label{eq:fuse}
\end{equation}
where $\lambda \in [0,1]$ is the weight chosen to trade off the relative importance of $E_{OmniEuc}$ and $E_{GIoU}$. Both terms are association costs normalised to $[0,1]$, where lower values indicate better matches. In this work $\lambda$ is a single weight per dataset, tracker, and detection source, optimised by grid search on the training split.
The overall pipeline flow of our system is shown in Fig.~\ref{fig:result_plot} (left), which also more clearly highlights the differences from existing SORT variants. 

\section{Data and Experiment Setup}
\label{sec:dataset}
We constructed OmniSmall from omnidirectional wildlife footage. We recorded eight clips of natural flying behaviour using \textit{QooCam 8K Enterprise}\footnote{\href{https://www.kandaovr.com/qoocam-8k-enterprise}{Kandao Technology: QooCam 8K Enterprise}} (7680$\times$3840, 30 FPS, 1,551 frames) and \textit{Ricoh THETA X}\footnote{\href{https://support.ricoh360.com/manuals/theta-x}{Ricoh THETA X}} (5760$\times$2880, 10 FPS, 900 frames) cameras. Both cameras produce equirectangular frames with $360^{\circ}$ horizontal coverage and hence a left-right wrap-around seam. A selected BBC Earth documentary clip\footnote{\href{https://www.xrportal.io/animals/red-kite-bird-feeding-frenzy-bbc-earth-unplugged/}{Red Kite Bird Feeding Frenzy | BBC Earth Unplugged}} (5120$\times$2560, 100 frames) was also included as a dense-flock stress test. OmniSmall contains 2,551 frames with manually annotated bounding boxes and identity labels for visible animals. Labels were assigned only while targets were visible; temporary exits therefore produce gaps and re-entries in trajectories.

Evaluation was also performed on JRDB~\cite{surround_2021,jrdb-2023}, a panoramic pedestrian-tracking benchmark recorded with a multi-camera rig on a mobile robot, stitched into a cylindrical panorama. Both datasets wrap horizontally and therefore contain a left-right seam. The JRDB annotations include a visibility or occlusion status for each object. Objects labelled as fully or partially visible were treated as ground-truth targets, while invisible or fully occluded objects were ignored. Table~\ref{tab:dataset_statistics} compares OmniSmall with JRDB. Small boxes were defined as those with area below $32^2$ pixels. 87.12\% of OmniSmall boxes satisfy this small-object criterion~\cite{tiny_review_2024}, against 5.24\% in JRDB.

For object detection, a single-class YOLOX detector~\cite{yolox_2021} was trained separately on each dataset using tiled crops; identity labels were not used for detector training. On JRDB, we follow the public training and validation splits; on OmniSmall, a chronological frame-level training-test split with a 2:8 ratio prevented training on future frames from the evaluation segment. This is to simulate the practical deployment in a remote field. High-resolution frames were processed in overlapping tiles: $1024 \times 1024$ with 200-pixel overlap for OmniSmall and $480 \times 480$ with 100-pixel overlap for JRDB. Tile-level detections were mapped back to full-frame coordinates and linear Soft-NMS~\cite{softnms_2017} was applied with an IoU threshold of $0.45$ and a confidence threshold of $0.10$. All trackers were evaluated on the same set of detections.

\begin{table}
\centering
\vspace{-2em}
\caption{Dataset statistics for OmniSmall and JRDB.}
\vspace{-.2cm}
\label{tab:dataset_statistics}
\setlength{\tabcolsep}{4pt}
\renewcommand{\arraystretch}{1.15}
\footnotesize
\begin{tabular}{lcc}
  \hline
  Statistic & OmniSmall & JRDB \\
  \hline
  Sequences & 9 & 54 \\
  Frames & 2,551 & 55,554 \\
  Tracks & 199 & 3,292 \\
  Annotated boxes & 20,771 & 1,256,363 \\
  Mean boxes per frame & 8.14 & 22.62 \\
  Mean box size (pixels) & $27.02 \times 16.24$ & $78.02 \times 169.55$ \\
  Mean box area / frame area (\%) & 0.0063 & 1.0754 \\
  Small boxes (\%) & 87.12 & 5.24 \\
  \hline
\end{tabular}
\vspace{-.6cm}
\end{table}

\section{Results and Discussion}
\label{sec:results}
\begin{table*}[t]
\caption{Performance comparison on JRDB and OmniSmall datasets with \textbf{ground-truth} detections. Best scores in each column are shown in \textbf{bold} and the second best is \underline{underlined}.}
\vspace{-.2cm}
\centering
\label{tab:jrdb_omnismall_comparison_gt}
\resizebox{\textwidth}{!}
{%
\begin{tabular}{lcccccccccccc}
\hline
\multicolumn{1}{c}{\multirow{2}{*}{\textbf{Method}}} &
  \multicolumn{6}{c}{JRDB-panorama} & \multicolumn{6}{c}{OmniSmall-omnidirectional} \\
\cmidrule(lr){2-7} \cmidrule(lr){8-13}
  & \multicolumn{1}{l}{$\lambda$} & \multicolumn{1}{l}{HOTA $\uparrow$} & \multicolumn{1}{l}{MOTA $\uparrow$} & \multicolumn{1}{l}{IDF1 $\uparrow$} & \multicolumn{1}{l}{IDSw $\downarrow$} & \multicolumn{1}{l}{FPS $\uparrow$} &
    \multicolumn{1}{l}{$\lambda$} & \multicolumn{1}{l}{HOTA $\uparrow$} & \multicolumn{1}{l}{MOTA $\uparrow$} & \multicolumn{1}{l}{IDF1 $\uparrow$} & \multicolumn{1}{l}{IDSw $\downarrow$} & \multicolumn{1}{l}{FPS $\uparrow$} \\
\cmidrule(lr){1-1} \cmidrule(lr){2-7} \cmidrule(lr){8-13}
\centering SORT~\cite{sort-2016} &
-- & 62.43 & 95.80 & 55.22 & 13819 & \underline{446} &
-- & 66.98 & 86.02 & 80.31 & 559 & 836 \\
\centering ByteTrack~\cite{bytetrack-2022} &
-- & 65.34 & 96.07 & \textbf{60.49} & 12856 & \textbf{575} &
-- & 71.84 & 90.88 & 90.12 & 94 & \textbf{1459} \\
\centering OCSORT~\cite{ocsort-2023} &
-- & 66.24 & \underline{96.14} & 56.20 & 13470 & 380 &
-- & 85.61 & 89.75 & 83.99 & 382 & 670 \\
\centering HybridSORT~\cite{hybridrsort_2024} &
-- & 66.80 & 96.00 & 56.86 & 11345 & 205 &
-- & 64.98 & 74.97 & 60.61 & 1247 & 273 \\
\centering OmniSORT + $E_{fuse}$ (ours) &
0.3 & \textbf{67.69} & 95.82 & \underline{58.44} & \textbf{8968} & 402 &
0.7 & \underline{94.10} & \textbf{99.24} & \underline{94.00} & \textbf{42} & \underline{1032} \\
\centering OmniOCSORT + $E_{fuse}$ (ours) &
0.2 & \underline{67.01} & \textbf{96.22} & 57.84 & \underline{9415} & 311 &
0.7 & \textbf{94.12} & \underline{99.16} & \textbf{94.16} & \underline{62} & 716 \\
\hline
\end{tabular}%
}
\vspace{-1.em}
\end{table*}
\begin{table*}[t]
\centering
\caption{Performance comparison on JRDB and OmniSmall datasets with \textbf{YOLOX} detections. Best scores in each column are shown in \textbf{bold} and the second best is \underline{underlined}.}
\vspace{-.2cm}
\label{tab:jrdb_omnismall_comparison_det}
\resizebox{\textwidth}{!}
{%
\begin{tabular}{lcccccccccccc}
\hline
\multicolumn{1}{c}{\multirow{2}{*}{\textbf{Method}}} &
  \multicolumn{6}{c}{JRDB-panorama} & \multicolumn{6}{c}{OmniSmall-omnidirectional} \\
  \cmidrule(lr){2-7} \cmidrule(lr){8-13}
  & \multicolumn{1}{l}{$\lambda$} & \multicolumn{1}{l}{HOTA $\uparrow$} & \multicolumn{1}{l}{MOTA $\uparrow$} & \multicolumn{1}{l}{IDF1 $\uparrow$} & \multicolumn{1}{l}{IDSw $\downarrow$} & \multicolumn{1}{l}{FPS $\uparrow$} &
    \multicolumn{1}{l}{$\lambda$} & \multicolumn{1}{l}{HOTA $\uparrow$} & \multicolumn{1}{l}{MOTA $\uparrow$} & \multicolumn{1}{l}{IDF1 $\uparrow$} & \multicolumn{1}{l}{IDSw $\downarrow$} & \multicolumn{1}{l}{FPS $\uparrow$} \\
\cmidrule(lr){1-1} \cmidrule(lr){2-7} \cmidrule(lr){8-13}
\centering SORT~\cite{sort-2016} &
-- & 26.45 & 27.94 & 25.94 & 22563 & \underline{332} &
-- & 39.04 & 20.31 & 45.88 & 117 & \underline{1152} \\
\centering ByteTrack~\cite{bytetrack-2022} &
-- & \textbf{31.92} & \textbf{46.02} & \textbf{35.45} & \textbf{8965} & \textbf{719} &
-- & 30.29 & 17.11 & 32.59 & \textbf{29} & \textbf{3175} \\
\centering OCSORT~\cite{ocsort-2023} &
-- & 26.45 & 27.42 & 25.85 & 22966 & 252 &
-- & 39.81 & 22.14 & 46.76 & 127 & 863 \\
\centering HybridSORT~\cite{hybridrsort_2024} &
-- & \underline{29.53} & \underline{35.83} & \underline{30.16} & \underline{16663} & 174 &
-- & 37.75 & 22.45 & 44.19 & 152 & 518 \\
\centering OmniSORT + $E_{fuse}$ (ours) &
0.2 & 25.35 & 27.94 & 24.35 & 22594 & 320 &
0.7 & \underline{41.11} & \underline{25.42} & \underline{49.33} & 137 & 1122 \\
\centering OmniOCSORT + $E_{fuse}$ (ours) &
0.2 & 27.70 & 35.20 & 27.58 & 18209 & 260 &
0.7 & \textbf{41.76} & \textbf{26.97} & \textbf{50.72} & \underline{108} & 862 \\
\hline
\end{tabular}%
}
\vspace{-1.em}
\end{table*}
For a fair comparison, we excluded trackers that employ appearance-based features and focused on TBD baselines. All baselines were run at published defaults. We first compared performance across trackers using ground-truth detections to isolate the tracking-stage contribution from detector noise. Table~\ref{tab:jrdb_omnismall_comparison_gt} shows that the proposed seam-aware association was most beneficial on OmniSmall, where targets are small and overlap-based association is least reliable. With ground-truth detections, OmniOCSORT improved over OCSORT on OmniSmall by +8.51 HOTA, +9.41 MOTA, and +10.17 IDF1, while IDSw fell from 382 to 62. OmniSORT improved over SORT by +27.12 HOTA and +13.22 MOTA. On JRDB, both seam-aware variants remained competitive: OmniSORT increased HOTA from 62.43 to 67.69 and OmniOCSORT increased HOTA from 66.24 to 67.01 (Fig.~\ref{fig:result_plot}, right). This contrast supports the intended operating regime of the method: small, fast targets in seam-affected omnidirectional footage, where overlap-only association is less reliable.

Performance with YOLOX detections (Table~\ref{tab:jrdb_omnismall_comparison_det}) was lower than with GT detections, as expected: missed detections break trajectories into new identities and damage all metrics. OCSORT's observation-centric re-update recovers tracks disrupted by missed detections, making OmniOCSORT the more robust variant under detector noise. OmniSORT fell slightly below the SORT baseline on JRDB (25.35 vs 26.45 HOTA): with noisy pedestrian-scale boxes a centre-distance cost can accept a false positive that non-overlap would reject, which suggests $E_{fuse}$ is better suited to small objects, as in OmniSmall.

On OmniSmall, OmniOCSORT with $E_{fuse}$ achieved the best performance, and its IDSw of 108 was second only to ByteTrack (29). On JRDB, where detections were dense and noisy, the score-aware ByteTrack achieved the best results, with HybridSORT second; OmniOCSORT improved over the SORT and OCSORT baselines (HOTA 27.70 vs 26.45) but did not match the dedicated two-stage score association of ByteTrack and HybridSORT. This indicates that under noisy pedestrian-scale detections the dominant signal remains dependent on detection-confidence handling and the multi-stage association, which our single-stage variants do not replicate.

We measured tracking-stage FPS on a single CPU core (AMD Ryzen 9 3900 12-Core Processor), since the proposed components affect only association and motion update; the tiled detector remains the dominant cost of the full pipeline. Under this protocol, OmniSORT with $E_{fuse}$ ran at 320 / 1122 FPS and OmniOCSORT with $E_{fuse}$ ran at 260 / 862 FPS on JRDB/OmniSmall with YOLOX detections. The seam-aware variants therefore remained within the same CPU-only tracking regime as the SORT-based baselines.

\vspace{-.1em}
\section{Ablation Study}
\label{sec:ablation}
To isolate the contribution of each proposed component, the ablation study was conducted on OmniSmall using ground-truth detections, incrementally building on the SORT and OCSORT baselines. The results are reported in Table~\ref{tab:ablation}.

SAMM alone reduced HOTA from 66.98 to 20.05 for SORT and from 85.61 to 23.67 for OCSORT. The velocity-change test is relative, $|\dot{u}_{f+1}-\dot{u}_{f}|/\max(|\dot{u}_{f}|,\epsilon)$, so for slow or near-stationary tracks the denominator saturates at $\epsilon$ and the wrap is applied beyond genuine seam events; in isolation this perturbs the velocity state while IoU-based association still requires overlap, and tracks are lost. The fused cost tolerates these perturbations because association no longer depends on overlap, so the combined configuration recovers and exceeds the baseline. 

The fusion weight $\lambda$ was swept in steps of $0.1$ across $[0,1]$. Both trackers achieved the highest HOTA at $\lambda=0.7$, while both endpoint configurations yielded lower HOTA. The single-term endpoints are complementary rather than redundant: OmniEuc alone is stronger for SORT (73.75 vs 60.53 HOTA) while GIoU alone is stronger for OCSORT (65.71 vs 52.23), and both fall well below the fused setting at $\lambda=0.7$ (94.10 and 94.12). Statistical significance was assessed with paired per-sequence exact sign-flip tests against the corresponding baselines. Summarising the three per-metric tests by their average $p$-value across HOTA, MOTA, and IDF1, the selected $\lambda=0.7$ configurations yielded $\bar{p}=0.0039$ for both trackers.
 
\begin{table}[t]
\centering
\caption{Component ablation and $\lambda$ sensitivity on \textbf{OmniSmall with GT detections}. The upper block of each tracker compares individual component substitutions; the lower block sweeps the representative fusion weight $\lambda$ in $E_{fuse}$. The $\bar{p}$ column reports the average $p$-value across the HOTA, MOTA, and IDF1 paired tests against the baseline ($0.0039$ is the exact-test floor of $\bar{p}$ at $n_{\mathrm{seq}}{=}9$). All configurations below the \textit{+ SAMM} row include SAMM. Best per tracker in \textbf{bold}.}

\label{tab:ablation}
\setlength{\tabcolsep}{4pt}
\renewcommand{\arraystretch}{1.2}
\footnotesize
\resizebox{\columnwidth}{!}
{%
\begin{tabular}{@{}llcccc@{}}
\hline
Tracker & Configuration & HOTA & MOTA & IDF1 & $\bar{p}$ \\
\hline
\multirow{4}{*}{SORT}
& baseline (IoU)   & 66.98 & 86.02 & 80.31 & --- \\
& + SAMM           & 20.05 & 18.29 & 14.20 & 0.0052 \\
& \quad + GIoU ($\lambda=0.0$)    & 60.53 & 69.20 & 56.14 & 0.0938 \\
& \quad + OmniEuc ($\lambda=1.0$) & 73.75 & 92.65 & 68.73 & 0.0260 \\
\hline
\multirow{5}{*}{
\shortstack[l]{OmniSORT \\ \quad + $E_{fuse}$}
}
& $\lambda=0.1$    & 69.86 & 78.58 & 66.04 & 0.2526 \\
& $\lambda=0.3$    & 85.33 & 90.52 & 83.68 & 0.0221 \\
& $\lambda=0.5$    & 91.07 & 94.83 & 90.67 & 0.0039 \\
& $\boldsymbol{\lambda=0.7}$ & \textbf{94.10} & \textbf{99.24} & \textbf{94.00} & 0.0039 \\
& $\lambda=0.9$    & 92.67 & 98.92 & 92.48 & 0.0039 \\
\hline\hline
\multirow{4}{*}{OCSORT}
& baseline (IoU)   & 85.61 & 89.75 & 83.99 & --- \\
& + SAMM           & 23.67 & 24.82 & 18.16 & 0.0052 \\
& \quad + GIoU ($\lambda=0.0$)    & 65.71 & 76.17 & 60.67 & 0.0104 \\
& \quad + OmniEuc ($\lambda=1.0$) & 52.23 & 71.03 & 36.29 & 0.1888 \\
\hline
\multirow{5}{*}{
\shortstack[l]{OmniOCSORT \\ \quad + $E_{fuse}$}
}
& $\lambda=0.1$    & 73.45 & 82.27 & 69.29 & 0.0078 \\
& $\lambda=0.3$    & 87.25 & 91.88 & 86.01 & 0.0078 \\
& $\lambda=0.5$    & 90.96 & 94.82 & 90.73 & 0.0039 \\
& $\boldsymbol{\lambda=0.7}$ & \textbf{94.12} & \textbf{99.16} & \textbf{94.16} & 0.0039 \\
& $\lambda=0.9$    & 66.78 & 84.96 & 56.02 & 0.0846 \\
\hline
\end{tabular}%
}
\end{table}

\vspace{-.1em}
\section{Conclusion}
\label{sec:conclusion}
We presented OmniSORT and OmniOCSORT, two lightweight SORT-based trackers re-engineered for deployable field studies using omnidirectional cameras. By incorporating SAMM and \Efuse, our approach improved identity preservation for small, fast targets in seam-affected omnidirectional footage while maintaining execution speed. Experiments showed large identity-aware gains on OmniSmall, especially when association quality is isolated using ground-truth detections, and competitive but mixed transfer to JRDB. These results suggest that the seam-aware motion model and composite association cost together target geometry- and scale-specific failure modes of IoU-based association that off-the-shelf SORT variants leave unaddressed, and OmniSmall provides a dedicated benchmark for studying small-object tracking in omnidirectional footage, though its scale limits the power of the per-sequence significance tests. Future work will strengthen the detector and integrate detection confidence into the seam-aware association, explore an adaptive fusion weight $\lambda$ in place of the current per-setting grid search, and narrow the remaining gap to score-aware trackers while preserving the appearance-free, CPU-only design intended for low-cost field deployment.

\vspace{-0.5em}
\bibliographystyle{IEEEtran}
\bibliography{refs}
\label{sec:refs}
\end{document}